%% file: main.tex
\documentclass[11pt,letterpaper]{article}

\usepackage[margin=1in]{geometry}
\usepackage[T1]{fontenc}
\usepackage[utf8]{inputenc}
\usepackage{lmodern}
\usepackage{microtype}
\usepackage{amsmath,amssymb,amsthm,mathtools}
\usepackage{graphicx}
\usepackage{booktabs}
\usepackage{array,tabularx}
\usepackage{enumitem}
\usepackage{caption}
\usepackage[numbers,sort&compress]{natbib}
\usepackage[colorlinks=true,citecolor=blue,urlcolor=blue,linkcolor=blue]{hyperref}
\usepackage{cleveref}
\crefname{figure}{Figure}{Figures}
\Crefname{figure}{Figure}{Figures}
\crefname{table}{Table}{Tables}
\Crefname{table}{Table}{Tables}
\crefname{section}{Section}{Sections}
\Crefname{section}{Section}{Sections}
\crefname{equation}{Equation}{Equations}
\Crefname{equation}{Equation}{Equations}
\usepackage{placeins}
\usepackage{url}
\usepackage{float}

\usepackage{orcidlink}
\newcommand{\nc}{\newcommand}
\nc{\yckorcid}{\orcidlink{0000-0001-7057-2096}}
\nc{\dc}{\definecolor}
\dc{watpink}{RGB}{198, 0, 120}
\dc{skyblue}{RGB}{0, 102, 204}
\dc{jade}{rgb}{0.0, 0.66, 0.42}
\dc{gsublue}{RGB}{0, 57, 166}
\nc{\red}[1]{\textcolor{red}{#1}}
\nc{\blue}[1]{\textcolor{blue}{#1}}
\nc{\green}[1]{\textcolor{green}{#1}}
\nc{\purple}[1]{\textcolor{purple}{#1}}
\nc{\bI}{\bm{\mathrm{I}}}
\nc{\violet}[1]{\textcolor{violet}{#1}}
\nc{\wpink}[1]{\textcolor{watpink}{#1}}
\nc{\jade}[1]{\textcolor{jade}{#1}}
\nc{\gsublue}[1]{\textcolor{gsublue}{#1}}
\nc{\watyellow}[1]{\textcolor{watyellow}{#1}}
\nc{\yck}[1]{\wpink{[YCK: #1]}}
\nc{\aimal}[1]{\gsublue{[Aimal: #1]}}

\newcommand{\K}{\mathcal K}

\newcommand{\I}{\mathbb I}
\DeclareMathOperator{\Retrieve}{Retrieve}
\DeclareMathOperator{\Unique}{Unique}
\DeclareMathOperator{\simil}{sim}
\DeclareMathOperator{\Finish}{Finish}
\DeclareMathOperator{\New}{New}
\setlist[itemize]{leftmargin=*,nosep,topsep=4pt}
\title{\bfseries Adaptive Retrieval, Selective Answers:\\
Evaluating Conformal Factuality in Agentic RAG}

\author{
\begin{minipage}[t]{0.46\textwidth}
\centering
Muhammad Aimal Rehman\\[0.3em]
{\small Department of Mathematics \& Statistics}\\
{\small Georgia State University}\\
{\small Atlanta, GA, USA}\\[0.3em]
\texttt{\small mrehman3@gsu.edu}
\end{minipage}
\hspace{0.03\textwidth}
\begin{minipage}[t]{0.46\textwidth}
\centering
Chi-Kuang Yeh \yckorcid\\[0.3em]
{\small Department of Mathematics \& Statistics}\\
{\small Georgia State University}\\
{\small Atlanta, GA, USA}\\[0.3em]
\texttt{\small cyeh@gsu.edu}
\end{minipage}
}

\date{}

\begin{document}
\maketitle

\begin{abstract}
Retrieval-augmented generation (RAG) improves the grounding of language-model responses through external evidence, yet retrieved information alone does not ensure factual support for generated claims. In agentic RAG, evidence acquisition is further complicated by adaptive retrieval decisions, where an agent iteratively determines whether to retrieve additional information or terminate the search. We investigate how conformal factuality can be applied in this setting to improve the reliability of generated responses. We formalize the agent-driven retrieval process and characterize a query-aware relevance-scoring formulation that accounts for evidence collected across successive retrieval decisions. Building on this formulation, we evaluate post-generation conformal claim filtering using GPT-6 Luna and Llama 3.1 8B across HotpotQA, Natural Questions, and TriviaQA. Our experiments demonstrate that increasingly stringent conformal targets are consistently associated with higher observed response-level factual support, accompanied by reductions in claim retention and non-empty response coverage. The extent of these trade-offs varies across models and datasets, highlighting the importance of jointly evaluating factual reliability and answer availability. These findings establish an empirical foundation for studying conformal factuality in agent-driven retrieval systems and motivate further investigation into improving the reliability of their generated responses.
\end{abstract}

\section{Introduction}\label{sec:intro}
Retrieval-augmented generation (RAG) equips language models with external evidence at generation time \citep{lewis2020retrieval}. Yet access to retrieved passages does not ensure that every claim in a generated response is supported by those passages. Conformal factuality addresses a related, post-generation question: given a set of generated claims and a calibrated selection rule, which claims should be retained? Earlier work developed conformal factuality for language-model outputs and retrieval-augmented settings \citep{mohri2024factuality,feng2025conformal}; our previous study investigated the approach in multi-hop RAG \citep{rehman2026multihop}.

Agentic retrieval makes this question especially relevant because the evidence set is constructed through decisions rather than a fixed sequence of searches. After an initial retrieval, a language-model policy may issue a follow-up query based on the evidence already available or terminate retrieval. The resulting collection of passages therefore depends on the path taken by the agent. Conformal filtering is applied only after this process has ended and a response has been generated, but the evidence collected along the way determines the scores used to select the response's claims. Understanding this separation between \emph{adaptive evidence acquisition} and \emph{post-generation factuality selection} is central to evaluating the system.

The retrieval trajectory also matters to the construction of the claim score itself. In our evaluated agentic pipeline, each stored passage carries a retrieval similarity associated with the query that first returned it. For a passage found through a follow-up query, that query need not coincide with the user's original question. This differs from representing all retrieved passages as if they had been scored against one fixed question. A precise description of the scoring rule must therefore account for the queries and evidence accumulated during the agent's actions, without assuming that a decision to stop certifies the sufficiency of the evidence.

We investigate two empirical questions: how observed response-level factual support changes as the conformal target becomes more stringent after agent-controlled retrieval, and how much non-empty answer content remains under this filtering across models and QA datasets. Our contributions are a query-aware formal description of the evaluated agentic retrieval and relevance-scoring process; a controlled evaluation spanning two language models, three datasets, five conformal targets, and no-CP baselines; and an analysis that reports both full-support rates and the availability of retained answers. Together, these elements describe how conformal factuality behaves when the evidence underlying generated claims is selected adaptively, without treating the experiment as a head-to-head ranking of RAG architectures.

The remainder of the paper develops the relevant literature in \cref{sec:background}, specifies the pipeline and calibration in \cref{sec:math}, and presents the experiments and results in \cref{sec:setup,sec:results}. Interpretation and scope are discussed in \cref{sec:discussion,sec:limitations}, followed by the conclusion and detailed appendices.

\section{Background and Relation to Prior Work}\label{sec:background}
\paragraph{Retrieval-augmented and multi-hop generation.}
RAG conditions generation on passages retrieved from a document collection \citep{lewis2020retrieval}. For questions requiring several pieces of evidence, later searches may use information obtained at earlier steps. HotpotQA was designed to evaluate compositional, multi-hop question answering \citep{yang2018hotpotqa}, while KILT links knowledge-intensive tasks to Wikipedia evidence and provenance \citep{petroni2021kilt}. Our earlier multi-hop conformal-RAG experiments used iterative query refinement subject to configured retrieval and stopping conditions \citep{rehman2026multihop}. The resulting evidence may span multiple retrieval steps even when the overall search procedure follows a prescribed loop.

\paragraph{Agent-controlled evidence acquisition.}
Agent-oriented language-model systems can interleave decisions with external actions, as exemplified by ReAct \citep{yao2023react}. In the evaluated agentic procedure, the policy selects either another retrieval query or a finish action after examining a bounded view of previously retrieved evidence. This is operationally distinct from merely repeating query rewriting until a preset rule stops the loop. \Cref{tab:operational} compares the two procedures used in our research, including their follow-up query generation, stopping controls, and evidence accumulation. In both cases, answer generation and conformal claim filtering occur downstream of retrieval; the distinction concerns how the evidence is assembled.

\paragraph{Conformal factuality and claim relevance.}
Conformal factuality uses a calibration sample to choose a selection threshold for a specified factual-support criterion \citep{mohri2024factuality}. When a generated response is decomposed into atomic claims, a relevance score can order those claims for retention, while factual-support annotations determine which calibration examples satisfy the criterion. These serve different roles: retrieval-based scores can be computed for new responses, whereas support annotations are needed to calibrate and evaluate the selection procedure. In retrieval-augmented generation, \citet{feng2025conformal} studied scores based on products of retrieval and claim--passage similarities, followed by a maximum over passages. Our work retains this product/max structure while evaluating it on evidence acquired over a sequence of policy-selected queries.

\paragraph{Positioning of this study.}
Prior single-stage and multi-hop investigations establish a basis for post-generation conformal claim filtering \citep{feng2025conformal,rehman2026multihop}. What changes in the present setting is the process that determines which passages enter the score: a document discovered after a follow-up query carries a stored retrieval similarity for that query rather than necessarily for the original question. This motivates the query-indexed formulation in \cref{sec:math} and the evaluation of factual support alongside answer availability across agentic retrieval trajectories.

\begin{table}[!htbp]
\centering
\caption{Operational differences between the evaluated multi-hop and agentic retrieval procedures. Both use post-generation claim scoring and conformal filtering.}
\label{tab:operational}
\small
\setlength{\tabcolsep}{4pt}
\begin{tabularx}{\linewidth}{@{}p{0.18\linewidth}XX@{}}
\toprule
\textbf{Dimension} & \textbf{Multi-hop RAG} & \textbf{Agentic RAG}\\
\midrule
Follow-up query & Rewritten from earlier-hop evidence & Generated by the policy upon \texttt{retrieve}\\
Retrieval control & Iteration within a hop budget & Explicit \texttt{retrieve}/\texttt{finish} decisions\\
Early termination & Hop limit or novel-evidence stopping rule & Policy \texttt{finish}, repeated-query safeguard, or budget\\
Evidence & Collected across retrieval hops & Collected from policy-selected searches\\
Response/CP stage & Generate, score, and filter claims & Same post-retrieval procedure\\
\bottomrule
\end{tabularx}
\end{table}

\section{Agentic Retrieval and Conformal Factuality}\label{sec:math}
The evaluated system has four stages. It first retrieves evidence for the original question; an agent policy then decides whether to issue further searches or finish. After the agent terminates, the model generates a response and decomposes it into atomic claims. Each claim receives a score from the accumulated evidence, and a threshold selected on calibration examples determines which claims are retained. The following formulation separates the agent's decisions from the post-generation filtering operation.

\subsection{Policy-driven retrieval with an adaptive stopping time}\label{sec:policy}
Let $x\in\mathcal X$ be a question and $\K$ a document corpus. We index \emph{retrieval operations}, rather than agent-policy calls, by $t$. The initial query is the original question, $q^{(1)}=x$. Given a top-$k$ retriever with similarity cutoff $\theta$, the initial document sequence is
\[
D_1=\Unique\!\left(\Retrieve(\K,q^{(1)};k,\theta)\right).
\]
After $t$ retrievals, let $Q_t=(q^{(1)},\ldots,q^{(t)})$ and let $D_t$ contain the distinct passages accumulated in retrieval order. Write the decision state and policy action as
\begin{equation}
H_t=\bigl(x,\Psi(D_t),Q_t,t,B-t\bigr),\qquad
 a_t=\pi_\omega(H_t)\in\{\Finish\}\cup\{\Retrieve(q):q\in\mathcal Q\}.
\label{eq:state}
\end{equation}
Here $\pi_\omega$ is the agent's policy, $B$ is the retrieval-operation budget, $\mathcal Q$ is the space of follow-up queries, and $\Psi(D_t)$ is the \emph{bounded evidence view} supplied to the policy, not necessarily the full accumulated corpus. Thus, the state in \cref{eq:state} distinguishes the evidence available to the policy from the full sequence of accumulated documents. If $a_t=\Retrieve(q^{(t+1)})$ is admissible, then
\begin{equation}
\begin{aligned}
E_{t+1}&=\Retrieve(\K,q^{(t+1)};k,\theta),\\
D_{t+1}&=D_t\mathbin{\oplus}\New(E_{t+1},D_t).
\end{aligned}
\label{eq:update}
\end{equation}
In \cref{eq:update}, $\New$ removes duplicates already present in $D_t$ and $\oplus$ appends previously unseen passages. The order matters because it can affect the evidence view available for the next policy decision.

Let $\tau(x)$ be the number of completed retrieval operations when the process terminates. Termination can result from a \texttt{finish} decision, the retrieval or decision-step limit, or safeguards against repeated queries; it need not be a calibrated confidence event. With an initial retrieval and bounded operations,
\[
1\leq\tau(x)\leq B,\qquad D_\pi(x):=D_{\tau(x)}.
\]
The retriever performs one initial retrieval before the policy makes follow-up decisions. This formulation permits different questions to have different retrieval trajectories without assuming independence between retrieval stages.

\subsection{Generation, claim decomposition, and relevance scoring}\label{sec:relevance}
After agent termination, the model generates a response and the claim decomposer extracts its atomic statements:
\[
\hat y=G\bigl(x,D_\pi(x)\bigr),\qquad
C(x)=g(\hat y)=\{c_1,\ldots,c_{p(x)}\}.
\]
The number of extracted claims $p(x)$ varies across questions. Write $A(c)\in\{0,1\}$ for the factual-support indicator of claim $c$, where $A(c)=1$ means supported. Support annotations are used in calibration and evaluation, not to choose a threshold from a test response.

The evaluated relevance scorer follows the product-and-maximum structure of \citet{feng2025conformal}. A document $d$ first added at retrieval operation $j(d)$ carries a stored retrieval similarity $s_{\rm ret}(d)$, evaluated for the query $q^{(j(d))}$ that retrieved it. The \emph{implemented} agentic relevance score can be expressed as
\begin{equation}
R^{\rm agent}(c;x)=
\max\!\left\{0,\max_{d\in D_\pi(x)}
  s_{\rm ret}(d)\,\simil(c,d)\right\}.
\label{eq:relevance}
\end{equation}
The similarity $\simil(c,d)$ measures semantic alignment between claim and passage. When retrieval similarities are cosine similarities in the same embedding space, $s_{\rm ret}(d)=\simil(q^{(j(d))},d)$. Thus, in \cref{eq:relevance}, documents retrieved during follow-up steps are scored against their corresponding agent-generated retrieval queries, whereas documents from the initial retrieval are scored against the original question $x$. For example, an initial retrieval for question $x$ might identify an entity that the policy then uses in a more specific follow-up query $q^{(2)}$. A passage first found through $q^{(2)}$ carries the retrieval factor $\simil(q^{(2)},d)$, rather than necessarily $\simil(x,d)$. This contrasts with the original-question factor in the single-query expression of \citet{feng2025conformal}, although the product/max aggregation structure is unchanged.

For tie handling, let $\widetilde R(c)=R^{\rm agent}(c;x)+\varepsilon(c)$ denote the score passed to conformal filtering, including any configured tie-breaking perturbation. The \emph{recorded experiments} use the thresholded retained set
\begin{equation}
F_\lambda(C(x))=\{c\in C(x):\widetilde R(c)\geq\lambda\}.
\label{eq:actualfilter}
\end{equation}
The filtering rule in \cref{eq:actualfilter} is applied after response generation; its threshold does not control when the retrieval agent stops.

\subsection{Calibration and response-level factuality}\label{sec:cp}
For a calibration question $i$, let $C_i$ be the extracted claims and $r_{ij}=\widetilde R(c_{ij})$ their relevance scores. Put $A_{ij}=1$ if claim $c_{ij}$ is supported and zero otherwise. The response is fully supported at threshold $\lambda$ when every retained claim is supported:
\begin{equation}
B_i(\lambda)=\I\left[\forall c_{ij}\in F_\lambda(C_i),\ A_{ij}=1\right].
\label{eq:fullsupport}
\end{equation}
The event defined in \cref{eq:fullsupport} is satisfied by a response retaining no claims, by convention. A useful explicit nonconformity score for \emph{all-claims-supported} calibration is
\begin{equation}
V_i=\max_{j:A_{ij}=0}r_{ij},\qquad \max\varnothing=-\infty.
\label{eq:nonconformity}
\end{equation}
The score in \cref{eq:nonconformity} is the largest relevance score of any unsupported claim. If $V_{(1)}\leq\cdots\leq V_{(n)}$ are the ordered calibration scores, the split-conformal order-statistic threshold is
\begin{equation}
k_\alpha=\left\lceil (n+1)(1-\alpha)\right\rceil,\qquad
\hat\lambda_\alpha=\begin{cases}V_{(k_\alpha)},&k_\alpha\leq n,\\+\infty,&k_\alpha=n+1.\end{cases}
\label{eq:quantile}
\end{equation}
Here $n$ is the calibration sample size and $1-\alpha$ is the target level. The specific sample size and target levels used with \cref{eq:quantile} are reported in \cref{sec:setup}.

\section{Experimental Design}\label{sec:setup}
\subsection{Datasets and models}
We evaluate HotpotQA \citep{yang2018hotpotqa}, Natural Questions (NQ) \citep{kwiatkowski2019nq}, and TriviaQA \citep{joshi2017triviaqa}, using KILT-linked validation resources and Wikipedia provenance \citep{petroni2021kilt}. Each experiment uses a deterministic 60-question subset selected under its dataset-specific answer/provenance and extractable-Wikipedia-evidence eligibility rules. For NQ and TriviaQA, the questions selected for the GPT-6 experiments were reused in the corresponding Llama experiments, matching those evaluation subsets across models. These eligibility conditions define the evaluated population and should be retained when reproducing the study.

The two generation backbones are the GPT-6 Luna API model (\texttt{gpt-6-luna}) and \texttt{meta-llama/Llama-3.1-8B-Instruct}. Each model family supplies the generator, agent-policy decisions, claim decomposition, and factuality verification within its own experimental pipeline. The common retrieval stack uses local sentence-transformer embeddings (\texttt{all-MiniLM-L6-v2}), a FAISS index, top-$k=10$ documents per retrieval, similarity cutoff $\theta=0.3$, and initial retrieval enabled. The retrieval-operation budget is $B=3$ and the separate policy-decision budget is also three. Additional implementation details, including the policy's evidence-view length and the handling of malformed structured outputs in the Llama experiments, are recorded in Table~\ref{tab:setup} in Appendix~\ref{app:implementation}.

\subsection{Calibration and evaluation protocol}
For each of the six model--dataset configurations, we generate, decompose, score, and annotate 60 question-level examples. Each claim's recorded annotation belongs to $\{S,I,U,N\}$, where only $S$ counts as supported. The within-response factuality requirement is $a=1$, corresponding to full support of all retained claims as defined in \cref{eq:fullsupport} and in our earlier study \citep{rehman2026multihop}. A no-CP baseline retains all generated claims. For each of the five CP targets, the same 60 scored/annotated examples are repeatedly divided into 30 calibration and 30 test cases over 100 seeded splits (base seed 42). Only the conformal threshold varies across target conditions within a split. The 100 runs are \emph{repeated partitions of fixed generated outputs}, not 100 independent retrieval-and-generation runs.

The experimental matrix is
\[
\{\text{GPT-6 Luna},\text{Llama 3.1 8B}\}
\times\{\text{HotpotQA},\text{NQ},\text{TriviaQA}\}
\times\{\text{No CP},80,85,90,92.5,95\%\}.
\]
The five conformal targets $1-\alpha\in\{0.80,0.85,0.90,0.925,0.95\}$ correspond, for the $n=30$ calibration examples, to $k_\alpha\in\{25,27,28,29,30\}$ under \cref{eq:quantile}. All six experiments use the product-based relevance score with maximum aggregation.

\subsection{Reported metrics}\label{sec:metrics}
For test example $i$, write $C_i(\lambda)=F_\lambda(C_i)$ and $n_i(\lambda)=|C_i(\lambda)|$. Let $m=30$ denote the test split size. The response-level full-support fraction and non-empty response coverage are
\begin{equation}
\begin{aligned}
\mathrm{FS}(\lambda)&=\frac1m\sum_{i=1}^{m}\I\bigl[\forall c\in C_i(\lambda),\ A(c)=1\bigr],\\
\mathrm{NE}(\lambda)&=\frac1m\sum_{i=1}^{m}\I[n_i(\lambda)>0].
\end{aligned}
\label{eq:metrics}
\end{equation}
The quantities in \cref{eq:metrics} distinguish support across all filtered outputs from the fraction retaining at least one claim. The mean-per-response fraction of retained claims is
\[
\mathrm{Ret}(\lambda)=\frac1m\sum_{i=1}^{m}\frac{n_i(\lambda)}{|C_i|}.
\]
Claim retention averages the retained fraction separately for each response, whereas the supported-claim fraction divides the number of supported retained claims by the total number of retained claims within each test split. The summary tables average these split-level measurements over the 100 repetitions. To evaluate factual support only for answers containing at least one retained claim, we use the conditional full-support fraction
\begin{equation}
\mathrm{FS}_{\mid\mathrm{NE}}(\lambda)=
\frac{\sum_i\I[n_i(\lambda)>0]\,\I[\forall c\in C_i(\lambda),A(c)=1]}
{\sum_i\I[n_i(\lambda)>0]},
\label{eq:FScond}
\end{equation}
when the denominator is nonzero. Unlike overall full support in \cref{eq:metrics}, the conditional quantity in \cref{eq:FScond} excludes empty outputs. The no-CP baseline always has 100\% non-empty coverage and 100\% claim retention for the successfully decomposed examples. We report percentages averaged across the 100 calibration--test partitions; they should not be treated as independent sample replicates.

\section{Results}\label{sec:results}
We examine factual support together with the fraction of questions that still receive a non-empty response after filtering. This distinction is important for reading the reliability--coverage curves. Let $N$ be the event that at least one claim is retained and let $\mathcal S$ be the event that every retained claim is supported. Under the full-support definition, an empty set of retained claims satisfies $\mathcal S$ by convention, giving
\begin{equation}
\Pr(\mathcal S)=1-\Pr(N)+\Pr(N)\Pr(\mathcal S\mid N).
\label{eq:decomposition}
\end{equation}
Equation~\ref{eq:decomposition} separates empty outputs from non-empty, fully supported outputs. We therefore interpret overall full support together with non-empty response coverage and full support conditional on a non-empty response.

\begin{figure}[H]
\centering
\includegraphics[width=\linewidth]{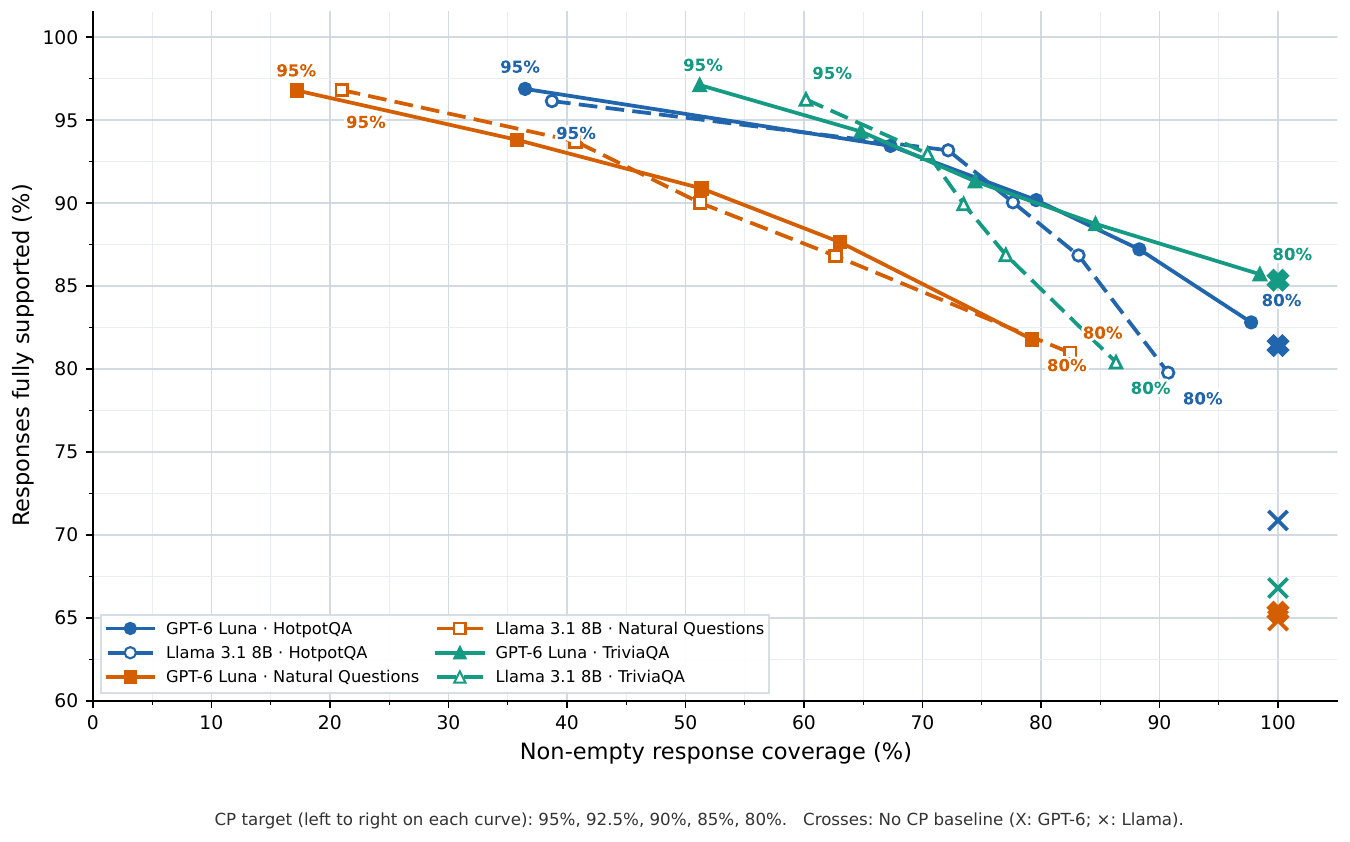}
\caption{Response-level full support versus non-empty answer coverage for GPT-6 Luna and Llama 3.1 8B on the three datasets. Curves show the five conformal targets; crosses at 100\% coverage represent no-CP baselines. Overall full support counts empty filtered responses as supported by convention.}
\label{fig:agentic_results}
\end{figure}

The reliability--coverage curves in \cref{fig:agentic_results}, measured by \cref{eq:metrics}, show a common direction across all six configurations: higher conformal targets accompany higher observed response-level full-support rates and fewer non-empty answers. The degree of selection, however, differs markedly among datasets and models. In particular, the Natural Questions curves reach substantially lower non-empty coverage than the TriviaQA curves at the same stringent targets. The plot thus reveals variation in the cost of claim filtering rather than a single uniform trade-off across experimental settings.

\Cref{tab:95summary} reports the most stringent target and the complementary content-retention measures. Each entry is computed from the experimental summaries, not estimated from the plotted curves.
\begin{table}[H]
\centering
\caption{Agentic RAG using the relevance-scoring formulation in \cref{eq:relevance} at the 95\% CP target. All values are percentages averaged over 100 calibration--test partitions. Conditional full support is calculated among non-empty responses.}
\label{tab:95summary}
\footnotesize
\setlength{\tabcolsep}{4pt}
\begin{tabular}{@{}llrrrr@{}}
\toprule
\textbf{Model} & \textbf{Dataset} & \shortstack{\textbf{Non-empty}\\\textbf{coverage}} & \shortstack{\textbf{Claims}\\\textbf{retained}} & \shortstack{\textbf{Supported}\\\textbf{claims}} & \shortstack{\textbf{Full support}\\\textbf{$\mid$ non-empty}}\\
\midrule
GPT-6 Luna & HotpotQA & 36.47 & 23.57 & 97.14 & 95.23\\
& NQ & 17.23 & 10.72 & 90.16 & 89.14\\
& TriviaQA & 51.20 & 43.16 & 96.59 & 95.56\\
\midrule
Llama 3.1 8B & HotpotQA & 38.73 & 27.06 & 97.33 & 93.26\\
& NQ & 21.00 & 11.77 & 96.27 & 91.47\\
& TriviaQA & 60.17 & 39.13 & 97.72 & 94.56\\
\bottomrule
\end{tabular}
\end{table}

At the 95\% target, non-empty coverage ranges from 17.23\% (GPT-6 Luna, NQ) to 60.17\% (Llama 3.1 8B, TriviaQA), while claim retention ranges from 10.72\% to 43.16\%. Both models exhibit the same ordering of the three datasets by non-empty coverage: NQ has the lowest coverage (17.23\% and 21.00\%), HotpotQA is intermediate (36.47\% and 38.73\%), and TriviaQA has the highest coverage (51.20\% and 60.17\%). This ordering is an observation about the selected model--dataset configurations, not evidence of a single cause for their differences. The supported-claim fractions at this target range from 90.16\% to 97.72\%, and full support conditional on non-empty output ranges from 89.14\% to 95.56\%, as defined in \cref{eq:FScond}.

The no-CP baselines provide another useful perspective. For GPT-6 Luna on NQ, full support conditional on non-empty output rises from 65.33\% without CP to 89.14\% at the 95\% target, while non-empty coverage falls from 100\% to 17.23\%. For Llama 3.1 8B on NQ, the corresponding conditional full-support rate rises from 64.83\% to 91.47\%, with coverage falling to 21.00\%. On TriviaQA, the 95\% target retains a greater proportion of non-empty responses (51.20\% for GPT-6 Luna and 60.17\% for Llama 3.1 8B) while reaching conditional full-support rates of 95.56\% and 94.56\%, respectively. The complete target sweeps and all no-CP baselines appear in \cref{tab:gpt6,tab:llama} in Appendix~\ref{app:tables}; these comparisons describe the observed joint changes in response support and availability, rather than attributing them solely to retrieval decisions.

\section{Discussion}\label{sec:discussion}
The results show that post-generation conformal filtering can be applied to claims produced after an agent has assembled its evidence through retrieval decisions. Across the evaluated configurations, more stringent targets are associated with progressively higher observed full support and more selective retained outputs. This pattern connects earlier conformal-factuality evaluations in single-stage and multi-hop RAG \citep{feng2025conformal,rehman2026multihop} to the agentic setting. The empirical contribution is not a change in the conformal objective but a characterization of its behavior when the evidence available for claim scoring arises from an adaptive retrieval trajectory.

Agentic evidence acquisition affects the input to the filtering stage in two related ways. First, the policy determines which passages enter the final evidence collection and when that collection stops growing. Second, a passage first obtained from a follow-up query contributes a stored retrieval similarity associated with that query, as made explicit in \cref{eq:relevance}. Consequently, the relevant scoring context is the assembled trajectory rather than only the original question and a single retrieval result. The conformal threshold acts after generation and does not itself direct the agent's searches. These distinctions explain why it is useful to represent retrieval decisions explicitly when studying downstream factuality filtering, without inferring that adaptivity alone caused the observed differences between experimental configurations.

The practical interpretation depends on both retained-answer availability and factual support. As the target becomes more stringent, a system may provide fewer answers containing claims, but the non-empty answers that remain exhibit higher observed conditional full-support rates in these experiments. The separation expressed in \cref{eq:decomposition} makes clear why overall full support, non-empty coverage, and conditional full support convey different information. Reporting them together permits comparisons of the observed trade-off across datasets and models, and makes the choice of a conformal target interpretable in terms of both supported content and response availability.

\section{Limitations}\label{sec:limitations}
Our evaluation uses controlled sets of 60 questions from each of three QA benchmarks. Dataset-specific answer, provenance, and retrievable-evidence requirements define the selected populations, so the results need not characterize the full benchmark distributions or settings using non-Wikipedia knowledge sources. Larger and more diverse question samples would provide a broader assessment.

The agent follows a bounded retrieval policy with at most three retrieval operations and three policy decisions. The embedding model, similarity threshold, and retrieval depth are also held fixed. We therefore do not evaluate longer agent trajectories, alternative policies, or the sensitivity of the observed results to different retrieval configurations.

The experiments cover GPT-6 Luna and Llama 3.1 8B and use one product/max retrieval-based relevance-scoring formulation. Other language-model families, such as Qwen and DeepSeek, and alternative claim-relevance scores are outside the present evaluation. Furthermore, the study does not implement a matched head-to-head comparison intended to isolate the effect of agentic retrieval relative to other RAG procedures; the findings concern conformal factuality within the evaluated agentic systems.

\section{Conclusion}\label{sec:conclusion}
We formulated bounded agentic RAG as a sequence of evidence-dependent retrieval decisions and evaluated post-generation conformal claim filtering over the evidence collected along those trajectories. Across six model--dataset configurations, increasing the conformal target was accompanied by higher observed full-support rates and lower claim retention and non-empty answer coverage. The differences between datasets and models show that the extent of this selection varies across settings. The central empirical lesson is that factual support and the availability of non-empty responses are complementary outcomes when evaluating conformal factuality after agent-driven evidence acquisition.

\section*{Code Availability}
The implementation of the Agentic RAG framework evaluated in this study is publicly available at \url{https://github.com/aimalrehman92/conformal-rag-agentic}.

\appendix
\section{Complete Target Sweeps and No-CP Baselines}\label{app:tables}
The following tables present the complete experimental results for GPT-6 Luna and Llama 3.1 8B across HotpotQA, Natural Questions, and TriviaQA. For each configuration, we report the no-CP baseline and five conformal targets (80, 85, 90, 92.5, and 95\%). The reported metrics are defined in \cref{sec:metrics}.

\input{Tables/agentic_gpt6_r0.tex}
\input{Tables/agentic_llama_r0.tex}

\section{Experimental Configuration Summary}\label{app:implementation}
\begin{table}[H]
\centering
\caption{Shared conditions across the six prioritized Agentic RAG experiments.}
\label{tab:setup}
\small
\begin{tabularx}{\linewidth}{@{}p{0.29\linewidth}X@{}}
\toprule
\textbf{Component} & \textbf{Specification}\\
\midrule
Experiment IDs & HotpotQA: E25 (GPT-6 Luna), E31 (Llama 3.1 8B); NQ: E27, E33; TriviaQA: E29, E35\\
Dataset & KILT-linked HotpotQA, Natural Questions, TriviaQA\\
Retrieval & Agentic \texttt{retrieve}/\texttt{finish}; initial retrieval enabled; top-10; similarity cutoff 0.3\\
Budgets & Maximum three retrieval operations and three policy decisions\\
Policy evidence view & Up to five passages per decision, each truncated to 1200 characters\\
Structured-output handling & Bounded retries for malformed agent/verifier outputs in the local Llama experiments\\
Embedding / index & \texttt{sentence-transformers/all-MiniLM-L6-v2} / FAISS\\
Score / filter & Product and max aggregation; experimental retention uses $\widetilde R(c)\geq\hat\lambda$\\
Conformal setting & Within-response factuality requirement $a=1$; targets 80, 85, 90, 92.5, 95\%\\
Protocol & 60 fixed questions per configuration; 30 calibration, 30 test; 100 seeded splits; seed 42\\
Baseline & No CP: original scored response with no claim removal\\
\bottomrule
\end{tabularx}
\end{table}

\newpage
\bibliographystyle{plainnat}
\bibliography{references}
\end{document}

%% file: Tables/agentic_gpt6_r0.tex
\begin{table}[H]
\centering
\caption{Agentic RAG with GPT-6 Luna using the relevance-scoring formulation in \cref{eq:relevance}: no-CP baseline and complete conformal target sweep (\%).}
\label{tab:gpt6}
\footnotesize
\setlength{\tabcolsep}{4pt}
\begin{tabular}{@{}llrrrr@{}}
\toprule
\textbf{Dataset} & \textbf{Target} & \textbf{Non-empty} & \textbf{Claims retained} & \textbf{Supported claims} & \shortstack{\textbf{Fully supported}\\\textbf{$\mid$ non-empty}} \\
\midrule
HotpotQA & \textbf{No CP} & 100.00 & 100.00 & 87.68 & 81.40 \\
 & 80\% & 97.73 & 92.57 & 88.63 & 82.66 \\
 & 85\% & 88.30 & 73.98 & 91.51 & 86.41 \\
 & 90\% & 79.60 & 62.28 & 93.33 & 88.82 \\
 & 92.5\% & 67.30 & 49.10 & 95.08 & 91.50 \\
 & 95\% & 36.47 & 23.57 & 97.14 & 95.23 \\
\midrule
NQ & \textbf{No CP} & 100.00 & 100.00 & 76.36 & 65.33 \\
 & 80\% & 79.27 & 62.95 & 86.78 & 78.06 \\
 & 85\% & 63.03 & 46.56 & 88.71 & 82.23 \\
 & 90\% & 51.37 & 36.04 & 89.03 & 84.53 \\
 & 92.5\% & 35.77 & 23.74 & 89.20 & 86.15 \\
 & 95\% & 17.23 & 10.72 & 90.16 & 89.14 \\
\midrule
TriviaQA & \textbf{No CP} & 100.00 & 100.00 & 87.63 & 85.33 \\
 & 80\% & 98.47 & 98.22 & 87.91 & 85.66 \\
 & 85\% & 84.60 & 82.51 & 89.22 & 87.59 \\
 & 90\% & 74.43 & 69.74 & 90.70 & 89.40 \\
 & 92.5\% & 64.80 & 58.34 & 93.16 & 91.99 \\
 & 95\% & 51.20 & 43.16 & 96.59 & 95.56 \\
\bottomrule
\end{tabular}
\end{table}

%% file: Tables/agentic_llama_r0.tex
\begin{table}[H]
\centering
\caption{Agentic RAG with Llama 3.1 8B using the relevance-scoring formulation in \cref{eq:relevance}: no-CP baseline and complete conformal target sweep (\%).}
\label{tab:llama}
\footnotesize
\setlength{\tabcolsep}{4pt}
\begin{tabular}{@{}llrrrr@{}}
\toprule
\textbf{Dataset} & \textbf{Target} & \textbf{Non-empty} & \textbf{Claims retained} & \textbf{Supported claims} & \shortstack{\textbf{Fully supported}\\\textbf{$\mid$ non-empty}} \\
\midrule
HotpotQA & \textbf{No CP} & 100.00 & 100.00 & 87.98 & 70.87 \\
 & 80\% & 90.73 & 84.19 & 91.66 & 78.26 \\
 & 85\% & 83.17 & 69.32 & 94.54 & 84.82 \\
 & 90\% & 77.63 & 60.33 & 95.61 & 87.88 \\
 & 92.5\% & 72.17 & 52.62 & 96.73 & 91.13 \\
 & 95\% & 38.73 & 27.06 & 97.33 & 93.26 \\
\midrule
NQ & \textbf{No CP} & 100.00 & 100.00 & 86.44 & 64.83 \\
 & 80\% & 82.47 & 62.21 & 88.75 & 77.54 \\
 & 85\% & 62.67 & 43.64 & 91.70 & 80.44 \\
 & 90\% & 51.23 & 34.42 & 93.05 & 82.16 \\
 & 92.5\% & 40.73 & 24.73 & 94.60 & 86.30 \\
 & 95\% & 21.00 & 11.77 & 96.27 & 91.47 \\
\midrule
TriviaQA & \textbf{No CP} & 100.00 & 100.00 & 86.38 & 66.80 \\
 & 80\% & 86.33 & 75.45 & 88.13 & 78.04 \\
 & 85\% & 77.03 & 61.78 & 91.58 & 83.69 \\
 & 90\% & 73.47 & 56.20 & 93.73 & 86.86 \\
 & 92.5\% & 70.43 & 50.02 & 95.82 & 90.51 \\
 & 95\% & 60.17 & 39.13 & 97.72 & 94.56 \\
\bottomrule
\end{tabular}
\end{table}